\documentclass[lettersize,journal]{IEEEtran}
\usepackage[]{caption}
\usepackage{subfigure}
\usepackage{amssymb}
\usepackage{multirow}
\usepackage{multicol}
\usepackage{color}
\usepackage{cite}
\usepackage{hyperref}
\usepackage{amsmath,amsfonts}
\usepackage{algorithmic}
\usepackage{array}
\usepackage{textcomp}
\usepackage{stfloats}
\usepackage{url}
\usepackage{verbatim}
\usepackage{graphicx}
\def\BibTeX{{\rm B\kern-.05em{\sc i\kern-.025em b}\kern-.08em
    T\kern-.1667em\lower.7ex\hbox{E}\kern-.125emX}}
\usepackage{balance}

\begin{document}
\title{AS-PD: An Arbitrary-Size Downsampling Framework for Point Clouds}
\author{Peng~Zhang,
	    Ruoyin~Xie,
	    Jinsheng~Sun,
	    Weiqing~Li,
	    and~Zhiyong~Su 
\IEEEcompsocitemizethanks{\IEEEcompsocthanksitem P. Zhang, R. Xie, J. Sun and Z. Su are with School of Automation, Nanjing University of Science and Technology, Nanjing 210094, China.\protect\\
E-mails: \{zhangpeng, xieruoyin\}@njust.edu.cn, jssun67@163.com, su@njust.edu.cn
\IEEEcompsocthanksitem W. Li is with School of Computer Science and Engineering, Nanjing University of Science and Technology, Nanjing 210094, China.\protect\\
E-mail: li\_weiqing@njust.edu.cn}
\thanks{Manuscript received April 19, 2005; revised August 26, 2015. (Corresponding author: Zhiyong Su.)}}

\markboth{Journal of \LaTeX\ Class Files,~Vol.~18, No.~9, September~2020}%
{AS-PD: An Arbitrary-size Downsampling Framework for Point Clouds}

\maketitle

\begin{abstract}
Point cloud downsampling is a crucial pre-processing operation to downsample points in order to unify data size and reduce computational cost, to name a few.
Recent research on point cloud downsampling has achieved great success which concentrates on learning to sample in a task-aware way. 
However, existing learnable samplers can not directly perform arbitrary-size downsampling, and assume the input size is fixed.
In this paper, we introduce the AS-PD, a novel task-aware sampling framework that directly downsamples point clouds to any smaller size based on a sample-to-refine strategy.
Given an input point cloud of arbitrary size, we first perform a task-agnostic pre-sampling on the input point cloud to a specified sample size.
Then, we obtain the sampled set by refining the pre-sampled set to make it task-aware, driven by downstream task losses.
The refinement is realized by adding each pre-sampled point with a small offset predicted by point-wise multi-layer perceptrons (MLPs).
With the density encoding and proper training scheme, the framework can learn to adaptively downsample point clouds of different input sizes to arbitrary sample sizes.
We evaluate sampled results for classification and registration tasks, respectively.
The proposed AS-PD surpasses the state-of-the-art method in terms of downstream performance.
Further experiments also show that our AS-PD exhibits better generality to unseen task models, implying that the proposed sampler is optimized to the task rather than a specified task model.
\end{abstract}

\begin{IEEEkeywords}
Point cloud, downsampling, arbitrary-size, point refinement
\end{IEEEkeywords}

\section{Introduction}
\IEEEPARstart{W}{ith} recent advances of 3D scanning techniques, point clouds tend to be the most popular form to represent 3D visual information. 
For most shape-level downstream tasks (e.g., classification \cite{liu-2021-cvpr,zhang-2022-neurcomp,qiu-2022-tmm,wang-2022-pr,deng-2022-cg}, and registration \cite{wang-2019-iccv,fu-2021-cvpr,qin-2022-cvpr,chen-2022-aaai}), the information involved in dense point clouds is always redundant.
When facing dense point clouds, it is hard for low-power devices or terminals to perform online inference and maintain fluent communication with other devices \cite{dovrat-2019-cvpr}.
In addition, the batch-wise parallel processing is widely used to encourage effective training and boost efficiency \cite{ioffe-2015-icml}, which requires input point clouds to be of uniform size.
Therefore, a downsampling operation, which will not change the data structure, is usually used as a pre-processing step to reduce or normalize the data size.

Existing sampling approaches can be roughly divided into two categories: heuristic methods and learning-based methods.
Heuristic methods are irrelevant to downstream tasks since they typically consider the low-level information and perform according to certain pre-determined rules.
Taking the widely used farthest point sampling (FPS) as an example, sampled points are selected in an iterative way: starting from a random initial point, the FPS samples the most distant point from the sampled set with regard to the rest points \cite{qi-2017-cvpr}.
Considering that heuristic methods are flexible in sample sizes and easily pluggable, they are widely used as a pre-processing step for point clouds.
However, since heuristic methods are task-agnostic, the sampled result is not optimal for different downstream tasks, causing serious performance degradation.

\begin{figure}[tbp] 
\centering 
\includegraphics[width=0.45\textwidth]{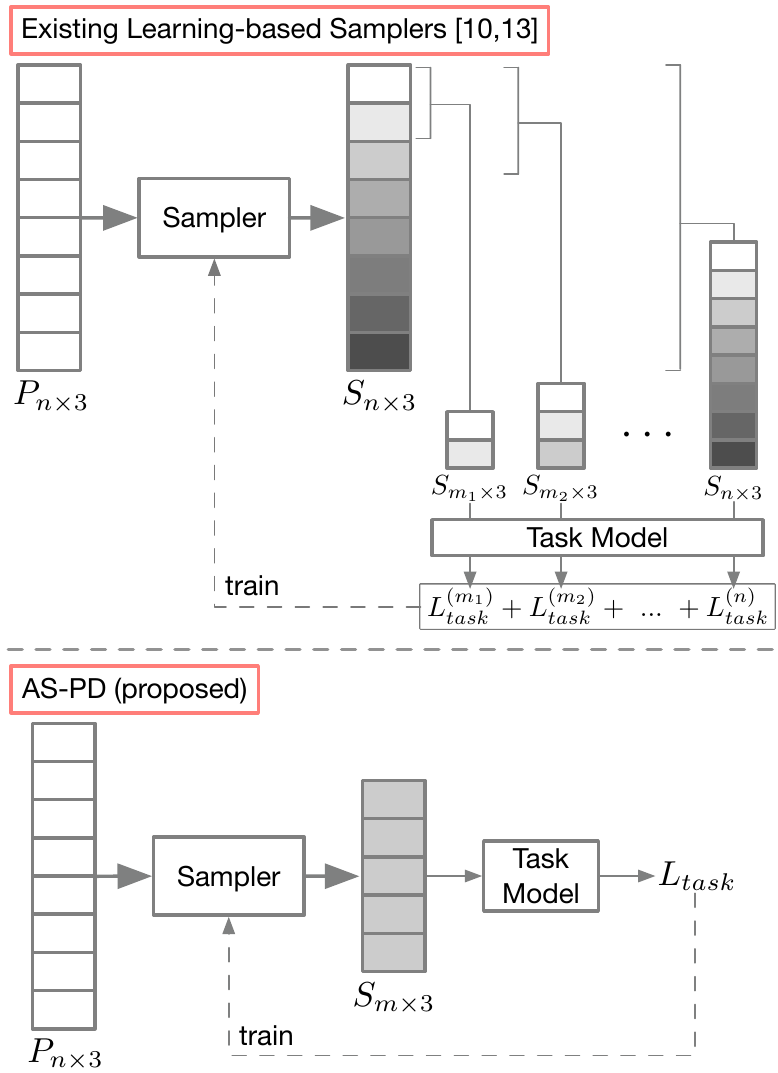} 
\caption{Illustration of sampling processes of existing learning-based samplers and the proposed AS-PD.
Existing approaches generate an intermediate result, consisting of points ordered by their importance to downstream tasks.
To make the generated points ordered, a progressive training scheme should define a set of task losses on several discrete sample sizes for one-time training.
In contrast, our AS-PD directly downsamples point clouds, which is trained in an end-to-end way.
} 
\label{Fig-introduction} 
\end{figure}

Learning-based methods \cite{dovrat-2019-cvpr}, \cite{lang-2020-cvpr} utilize neural networks to build a learnable sampler that is optimized with downstream task constraints.
Existing approaches take an indirect way to realize arbitrary-size downsampling.
As illustrated in Fig \ref{Fig-introduction}, existing learning-based samplers first generate a point set of the same size as the input.
Then, with a progressive optimization scheme, the generate points are reordered according to their importance to downstream tasks.
Finally, to obtain the sampled set of a given size, they select points according to their order in the generated set.
However, such an indirect manner still lacks flexibility.
On the one hand, existing approaches assume that the input point clouds are of the same size, which is impractical.
On the other hand, the indirect manner relies on a complicated optimization scheme.
As shown in Fig. \ref{Fig-introduction}, to make the generated points ordered, a set of task constraints are defined on several discrete sample sizes for one-time training.

To this end, we present a novel arbitrary-size downsampling framework, termed AS-PD, which takes advantages of both heuristic methods and learning-based methods.
We first take a flexible and uniform heuristic method (e.g., FPS) as a pre-sampler to perform arbitrary-size but task-agnostic downsampling on the input point cloud.
Then, a point-wise refiner is exploited to adaptively refine the initial sampled result, driven by the downstream task.
The proposed sample-to-refine pipeline naturally satisfies downsampling for arbitrary input size and sample size, since the pre-sampler is flexible and the subsequent refinement is point-wise.
Thus, the sampling process is straightforward and can be trained in an end-to-end way, as illustrated in Fig. \ref{Fig-introduction}.
In addition, since our sampled set is evolved from a uniform pre-sampled set rather than from the air, the sampled points are more likely to recover the details of the input shape.
In this way, we can prevent the sampled set from overfitting the task model instead of the task.

To sum up, our key contributions are three-fold:
\begin{itemize}
\item We propose a novel task-aware downsampling framework, termed AS-PD, which directly downsamples point clouds of different input sizes to arbitrary sample sizes.
\item The proposed AS-PD tends to preserve the details of input point clouds, which prevents the sampled set from overfitting the specific task model.
\item We conduct extensive experiments and ablation studies on downstream tasks including classification and registration. Our experiments indicate that the AS-PD outperforms the state-of-the-art method in many aspects, including downstream performance and generality, etc.
\end{itemize}

\section{Related Works} \label{sec:related_works}
\subsection{Point Cloud Downsampling in Neural Network}
Recently, several works concentrated on how to perform sampling inside the neural network for better feature learning.
Traditional sampling methods (e.g., FPS) are widely used in hierarchical structures to select representative points for feature aggregation \cite{qi-2017-nips}, \cite{li-2018-nips}, \cite{hu-2021-tpami}.
Considering traditional methods mainly rely on low-dimensional geometric features and can not adapt to different cases, better points-selecting methods are expected.

Yang et al. \cite{yang-2019-cvpr} propose a sampling operation, named Gumbel Subset Sampling (GSS), to replace FPS in the hierarchical feature embedding process.
With a discrete reparameterization trick, GSS produces a ``soft" continuous subset to approximate the discrete sampling during the training process.
Yan et al. \cite{yan-2020-cvpr} also propose a noise-robust adaptive sampling module to replace FPS in neural networks.
Nezhadarya et al. \cite{nezhadarya-2020-cvpr} define ``critical points" according to the feature value of each point, and sample those critical points.
Lin et al. \cite{lin-2022-tvcg} propose a task-aware sampling layer to reserve critical points for downstream tasks in a learnable way.

Note that the sampling methods mentioned above are all embedded in the network structure to improve the network performance.
While in this paper, we concentrate on the downsampling that is taken as an offline pre-processing step.

\begin{figure*}[!t] 
	\centering 
	\includegraphics[width=0.9\textwidth]{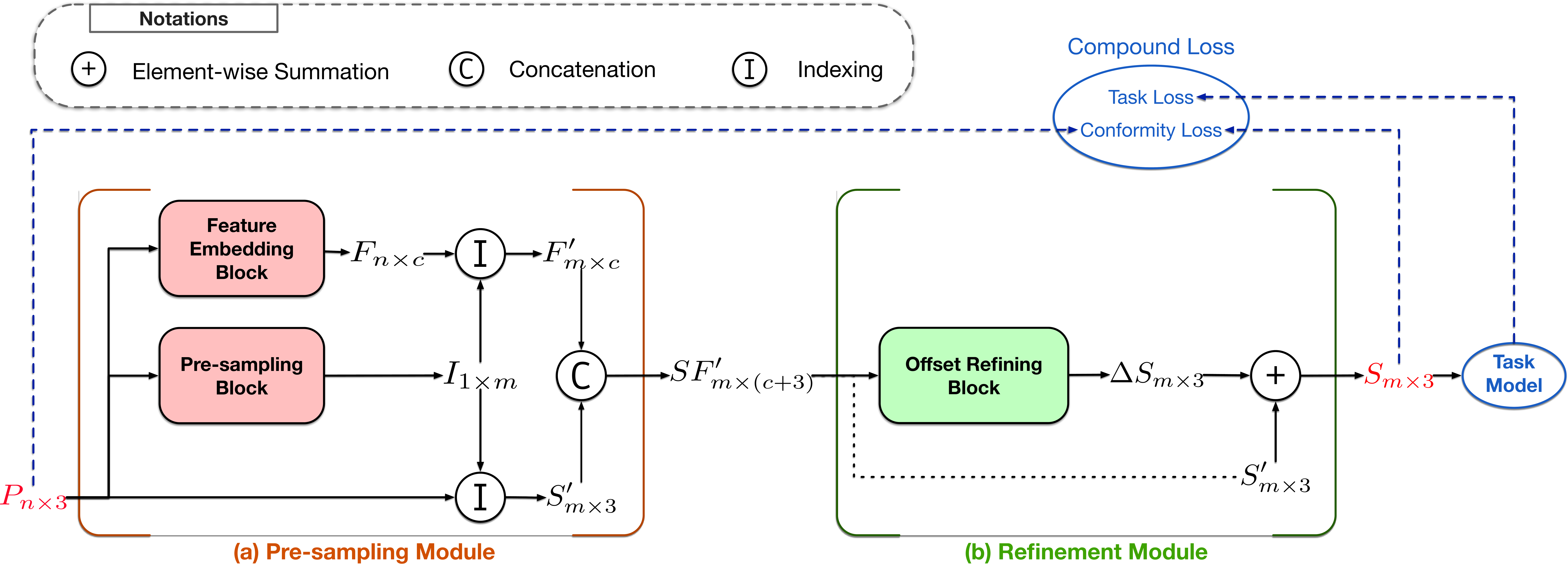} 
	\caption{Framework of the proposed AS-PD.
		Given a point set $P_{n\times 3}$ with $n$ points, a pre-sampling module is first utilized to give pre-sampled points with features $SF'_{m\times (c+3)}$, then the pre-sampled point set $S^{\prime}_{m\times 3}$ is refined by a refinement module.
		$P_{n\times 3}$ and $S_{m\times 3}$ represent the input and output of the framework, which are both colored in red.
		Ingredients related to loss are colored in blue and only occur during the training stage.} 
	\label{Fig-framework} 
\end{figure*}

\subsection{Point Cloud Downsampling as Pre-process}
Point cloud downsampling, as a pre-process procedure, can be roughly divided into two categories: traditional methods and learning-based methods.

Traditional methods typically originate from certain heuristic priors, so they are also called heuristic methods.
Farthest point sampling (FPS) and random sampling (RS) are two of the most popular heuristic methods in the literature \cite{qi-2017-cvpr,qi-2017-nips,li-2018-nips,wu-2019-cvpr,guo-2021-cvm,hu-2021-tpami}.
The RS takes downsampling as a random selection process, and therefore achieves the highest speed while inevitably losing some important information of the original data due to the randomness.
The FPS randomly starts from a point in a point cloud and selects the next point that is farthest away from the sampled point set.
Such an iterative sampling procedure ensures a good coverage of the original point cloud.
Following the FPS, there are many other methods pursuing a good coverage to avoid information loss, e.g., poisson disk sampling \cite{gamito-2009-tog} (PDS) and voxel sampling \cite{zhang-2020-eccv} (VS).
Although widely used in the literature, heuristic methods are task-agnostic and thus not optimal for different tasks.

Learning-based methods typically model the discrete sampling as a continuous point generation process driven by downstream tasks.
Dovrat et al. \cite{dovrat-2019-cvpr} take the lead in performing learnable sampling by generating the sampled points from a global latent code, named S-NET.
Hereafter, a few works refine S-NET based on its original structure \cite{lang-2020-cvpr,qian-2020-arxiv,wang-2021-icig,lin-2021-prai,wang-2022-tvc}.
However, the generative way is unfavorable for the arbitrary-size downsampling due to the fixed network structure.
Some other learning-based methods have attempted to realize arbitrary-size downsampling.
Dovrat et al. \cite{dovrat-2019-cvpr} propose a progressive version of the S-NET to indirectly realize the arbitrary-size downsampling, termed ProgressiveNet (PN).
Lang et al. \cite{lang-2020-cvpr} further improve the PN by inserting a soft projection process at the end of the network, termed SampleNet Progressive (SNP).
These progressive samplers generate an ordered point set of the same size as the input, and indirectly downsample points by selecting points according to their order in the generated set.

However, the indirect arbitrary-size downsampling approach still lacks flexibility and generality.
First, the input size is assumed to be fixed during downsampling.
Second, the output of the sampling network is not the sampled set, which obstructs effective end-to-end training.
At last, the sampled set is easily overfitted to the task model instead of the task, since the sampled points always miss a lot of information of the input.
Hence, in this paper, we design a novel direct arbitrary-size downsampling framework to solve the aforementioned drawbacks.

\subsection{Point Refinement Technique}
Point refinement techniques typically refine coarse point clouds by modifying the coordinates of each point.
It is widely used in the field of point cloud denoising \cite{zhang-2020-tvcg,rakotosaona-2020-cgf,metzer-2021-tog,chen-2022-ijcv} and upsampling \cite{qian-2021-tip,li-2021-cvpr,lin-2022-tvcg,li-2022-tvcg,zhao-2022-cvpr}.
The refinement procedure is usually driven by geometric constraints (e.g., Chamfer Distance \cite{fan-2017-cvpr} and Earth' mover distance \cite{rubner-2000-ijcv}), so as to reduce the geometric error between the input points and denoised (or upsampled) points.

Zhang et al. \cite{zhang-2020-tvcg} use neural networks to regress a displacement vector between the noisy point and the underlying surface for point cloud denoising.
Metzer et al. \cite{metzer-2021-tog} let a spatial refiner learn to refine a coarse source point set to a sharp target point set, and implicitly learn to consolidate the point cloud.
The spatial refinement is performed by adding coordinate offsets to the source point set.
Li et al. \cite{li-2022-tvcg} define a sparse point set representation which is better for dense point cloud recovery, term the point set self-embedding.
The self-embedded point set is obtained by spatially refining the pre-sampled subset.
To achieve point cloud upsampling, qian et al. \cite{qian-2021-tip} first densify the input point cloud by interpolation, and then refined the interpolated points by spatial refinement.
Lin et al. \cite{lin-2022-tvcg} design a task-aware sampling layer based on point refinement, to improve feature learning in hierarchical networks.

Though the point refinement technique shows great potential in many cases, its application in point cloud downsampling has received little attention.
Furthermore, the arbitrary-size challenge for point cloud downsampling hinders the practicality of existing learning-based samplers.
Therefore, we design a sample-to-refine strategy to address the inflexibility of existing learning-based approaches.

\section{Method} \label{sec:method}
\subsection{Overview}
In this paper, we design a novel task-aware downsampling framework, termed AS-PD, which directly downsamples point clouds to arbitrary sizes.
Given a point set $P=\{p_i\}_{i=1}^{n}$ with $n$ points, the AS-PD aims to sample a simplified point set $S=\{s_{j}\}_{j=1}^{m}$ with $m$ points, where $n$ and $m$ can be set to arbitrary numbers subjected to $n\textgreater m$.
Fig. \ref{Fig-framework} depicts the pipeline of the AS-PD, including (a) a pre-sampling module for feature embedding and pre-sampling, and (b) a refinement module for refining pre-sampled points.

The pre-sampling module extracts features $F_{n\times c}$ from the input points $P_{n\times 3}$, and generates a pre-sampled set $S^{\prime}_{m\times 3}$ for a given sample size $m$.
Then, the refinement module adaptively predicts the point-wise offset $\Delta S_{m\times 3}$ for each point in the pre-sampled set $S^{\prime}_{m\times 3}$ and gets the refined sampled set $S_{m\times 3}$.
With a compound loss and a two-stage training scheme, the refined set $S_{m\times 3}$ keeps uniform distributed on the underlying surface of point clouds while being task-aware across different sample sizes.

\begin{figure*}[!t] 
	\centering 
	\includegraphics[width=0.9\textwidth]{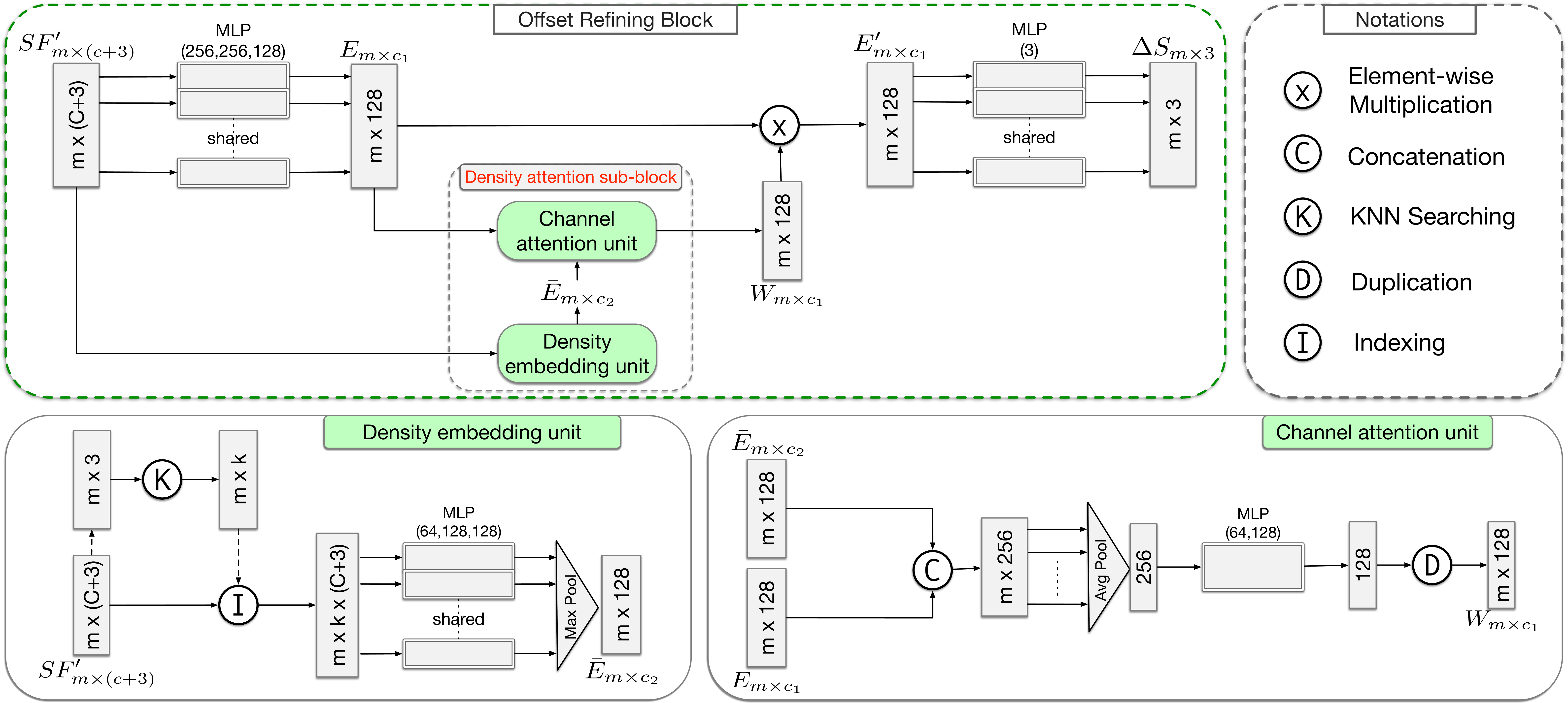} 
	\caption{Network architecture of the offset refining block.
			The upper two rectangles represent the network architecture of the offset refining block and notations for used flags, respectively. The offset refining block contains a density attention sub-block, consisting of a density embedding unit and a channel attention unit, whose details are illustrated in the lower two rectangles.}
	\label{Fig-refinement_module} 
\end{figure*}

\subsection{Pre-sampling Module} \label{pre-sampling}
The pre-sampling module takes in the input point set $P_{n\times 3}$ and outputs a pre-sampled set $S^{\prime}_{m\times 3}$ with features $F^{\prime}_{m\times c}$, where $n$ and $m$ denote the input size and sample size ($m \textless n$). 
It mainly consists of a feature embedding block and a pre-sampling block. 

The feature embedding block is exploited to extract high-dimensional features $F_{n\times c}$ of the input point set $P_{n\times 3}$.
As the part that directly handles input point clouds, the feature extractor should be able to deal with different input sizes.
In this paper, we take the widely used DGCNN \cite{wang-2019-tog} as the backbone for the feature embedding block, considering its excellent ability in learning local geometric features.
As a point-based method, DGCNN \cite{wang-2019-tog} can naturally process the input sets across different sizes, once the neighborhood size $k$ is properly set.
Note that other advanced point-based feature extractors (e.g., CurveNet \cite{xiang-2021-iccv} and PCT \cite{guo-2021-cvm}) could also be alternatives to this block, once meeting the requirement of the arbitrary-size input processing. 

The pre-sampling block is used to downsample the input point cloud to a given arbitrary size. 
The sampled result is described by the point indices $I_{1\times m}$, which indicate the positions of the sampled points in the input point cloud.
To realize arbitrary-size downsampling, the selected pre-sampler should be able to directly downsample point clouds to arbitrary sizes.
In this paper, we take the FPS as the pre-sampler for the pre-sampling block, considering its good coverage of the input point cloud.
It is noteworthy that other flexible heuristic methods (e.g., RS), which can downsample point cloud to any size, could also be alternatives to the pre-sampler. 

Finally, we can get the pre-sampled set $S^{\prime}_{m\times 3}$ and corresponding point features $F^{\prime}_{m\times c}$ based on the indices $I_{1\times m}$ of pre-sampled points.
The concatenation of $S^{\prime}_{m\times 3}$ and $F^{\prime}_{m\times c}$ constructs the output of the pre-sampling module $SF^{\prime}_{m\times(c+3)}$.

\subsection{Refinement Module} \label{refinement}
The refinement module takes in the concatenation of pre-sampled coordinates and features $SF^{\prime}_{m\times(c+3)}$, and outputs a task-aware refined set $S_{m\times 3}$ driven by downstream task losses.
The refinement is implemented by a carefully designed offset refining block, as illustrated in Fig. \ref{Fig-refinement_module}.

The offset refining block consists of point-wise MLPs and a density attention sub-block.
The point-wise MLPs guarantee the ability to deal with point sets of arbitrary size.
The density attention sub-block is used to adaptively adjust the point features $E_{m\times c_1}$ to make it aware to the sample size.
In detail, the density attention sub-block consists of a density embedding unit and a channel attention unit, whose network architectures are depicted in Fig. \ref{Fig-refinement_module}.

The density embedding unit implicitly extracts sample size cues from pre-sampled coordinates and features $SF^{\prime}_{m\times(c+3)}$, and embeds it into local features $\Bar{E}_{m\times c_2}$.
The way of capturing sample size cues is oriented from such observation:
the local density of the pre-sampled set $S^{\prime}_{m\times 3}$ changes with the sample size $m$.
Therefore, by implementing the feature extracting on the fixed $k$-NN region for each point in $S^{\prime}_{m\times 3}$, we can obtain local features related to the sample size.
Specifically, the density embedding unit first searches local $k$-NN neighbors for each point in the pre-sampled set $S^{\prime}_{m\times 3}$. 
Then, it applies a PointNet-like operation on each local region. 
Finally, it gets local features $\Bar{E}_{m\times c_2}$ for each point in the pre-sampled set $S^{\prime}_{m\times 3}$.

The channel attention unit takes in the concatenation of point features $E_{m\times c_1}$ and $\Bar{E}_{m\times c_2}$, which originate from the shared point-wise MLPs and the density embedding unit, respectively.
With the average pooling and MLPs, a channel-wise attention vector with length of $c_1$ ($c_1$=128) is obtained.
We construct the channel attention matrix $W_{m\times c_1}$ by duplicating the channel-wise attention vector $m$ times.
Then, the adjusted point features $E^{\prime}_{m\times c_1}$ can be achieved as follows:
\begin{equation}\label{eq-attention}
E^{\prime}_{m\times c_1} = E_{m\times c_1} \odot W_{m\times c_1},
\end{equation}
where $\odot$ denotes the Hadamard product.
After that, we use point-wise MLPs to project the adjusted point features to the Euclidean space, and get the point-wise offset value $\Delta S_{m\times 3}$.

Finally, by adding the offset value $\Delta S_{m\times 3}$ to the pre-sampled set $S^{\prime}_{m\times 3}$, the refined sampled set $S_{m\times 3}$ is obtained, as shown in Equation \ref{eq-offset}:
\begin{equation}\label{eq-offset}
S_{m\times 3} = S^{\prime}_{m\times 3} + \Delta S_{m\times 3}.
\end{equation}

\subsection{Loss Functions} \label{loss_function}
To train the sampler for specific downstream tasks in an end-to-end way, we utilize a compound loss function consisting of the task loss, conformity loss, and offset loss, as shown in Equation \ref{eq-overall_loss}:
\begin{equation}\label{eq-overall_loss}
L = \lambda L_{task}(S) + \alpha L_{conf}(P, S) + \beta L_{off}(S', S),
\end{equation}
where $\lambda$, $\alpha$ and $\beta$ are utilized to balance the three loss terms, respectively.

\subsubsection{Task Loss}
The task loss is defined on the refined sampled set $S_{m\times 3}$.
The design of task loss depends on downstream tasks. 
For example, a negative likelihood loss is appropriate for the classification task, while a rotation error loss is a typical alternative to the registration task. 

\subsubsection{Conformity Loss}
The conformity loss is defined between the input set $P_{n\times 3}$ and the refined sampled set $S_{m\times 3}$, which aims at encouraging refined points to be close to input points. 
We take the Chamfer Distance \cite{fan-2017-cvpr} between $P_{n\times 3}$ and $S_{m\times 3}$ as the conformity loss:
\begin{equation}\label{eq-conf_loss}
\begin{aligned}
L_{conf} &= \frac{1}{|S|} \sum_{s \in S} \min_{p\in P} \|s-p\|^2 + \frac{1}{|P|} \sum_{p \in P} \min_{s\in S} \|p-s\|^2, \\
\end{aligned}
\end{equation}
where $\|.\|$ represents the $L^2$ norm.
$|S|$ and $|P|$ represent the number of points in $S_{m\times 3}$ and $P_{n\times 3}$, respectively.

\subsubsection{Offset Loss}
The offset loss suppresses the long-range movement to guarantee that the refined set $S_{m\times 3}$ remains close to the underlying surface.
We define the offset loss as the average distance for moving a point:
\begin{equation}\label{eq-off_loss}
L_{off} = \frac{1}{|S|} \sum_{s^{\prime}, s \in (S^{\prime},S)} \|s-s'\|,
\end{equation}
where $|S|$ is the number of points in $S_{m\times 3}$, $\|.\|$ is the $L^2$ norm, and $(S^{\prime},S)$ represents a set of point pairs.
Each pair in $(S^{\prime},S)$ contains points before and after moving, respectively.

\subsection{Training Scheme} \label{method-training_scheme}
We take a two-stage way to train the AS-PD.
In the first stage, we cut-off the density attention sub-block of the offset refining block in the refinement module, and train the left part of the AS-PD with a relatively small sample size.
In the second stage, we load the pre-trained weights from the first stage, and take the density attention sub-block back.
After that, we retrain the AS-PD with sample sizes that varies for each batch of data.
Specifically, for each batch of data, the sizes of the input set and sampled set are both selected from the respective candidate sets.

\section{Experiment} \label{sec:experiment}
In this part, we conduct a series of quantitative and qualitative experiments on downstream tasks including classification and registration, in order to validate the effectiveness of our sampling framework.
Further experiments are conducted to justify the flexibility and generality of the sampling framework.
Finally, we do ablation studies on components in the pipeline of the proposed AS-PD, to validate their effectiveness.

\subsection{Experiment Setup} \label{exp-setup}
\textbf{General implementation details}.
In the pre-sampling module, we utilize the DGCNN \cite{wang-2019-tog} to embed features.
For the input of 1024 points, we set the neighborhood number to 40 for the local $k$-NN graph construction.
Other detailed structure follows the official setting for part segmentation.
Besides, the FPS is used to perform arbitrary-size and uniform pre-sampling.
In the refinement module, several MLP layers are used to build the offset refinement block.
The detailed architecture of the offset refining block has been clearly illustrated in Fig. \ref{Fig-refinement_module}.
Further task-specific implementation details can be found in the following subsections.

\textbf{Compared methods}.
We compare our technique with the most popular heuristic methods (i.e., RS and FPS \cite{qi-2017-cvpr}) as well as the learning-based methods (i.e., PN \cite{dovrat-2019-cvpr} and SNP \cite{lang-2020-cvpr}).
We choose PN and SNP as compared methods considering that these task-aware samplers target arbitrary-size sampling as we do.
Note that there exist other task-aware samplers, such as S-NET \cite{dovrat-2019-cvpr}, SampleNet \cite{lang-2020-cvpr} and Point Displacement Network \cite{lin-2022-tvcg}.
But they are either unable to perform arbitrary-size sampling, or designed for the different sampling perspective that takes the sampler as an embedded layer in neural networks.

\textbf{Evaluation metrics}.
We take the downstream task performance of the sampled set as the main metric to evaluate the sampler.
For the classification task, we use the classification accuracy (Acc) of the sampled set to assess the downstream task performance.
For the registration task, we take the mean rotation error (MRE) \cite{yuan-2018-3dv} to evaluate the downstream task performance.
For each pair of samples, the rotation error (RE) can be described as follows:
\begin{equation}\label{eq-RE}
\begin{aligned}
RE = 2 cos^{-1} (2\left<q_{pred}, q_{gt}\right>^2 - 1),
\end{aligned}
\end{equation}
where $\left<.,.\right>$ represents inner product, $q_{pred}$ and $q_{gt}$ represent the quaternions for the predicted rotation and the ground truth.
We measure the rotation in degrees (DEG) instead of radians.

We take the Hausdorff Distance \cite{huttenlocher-1993-tpami} (HD) as the metric to evaluate the conformity of the sampled set and the input set.
We do not take the Chamfer Distance as the conformity metric, since it is a part of our loss function.
A third-party metric such as HD is more objective.
Note that compared methods (i.e., PN \cite{dovrat-2019-cvpr} and SNP \cite{lang-2020-cvpr}) both introduce a matching process additionally, which is not involved in our pipeline.
The matching process improves the conformity to the input set, but prevents the end-to-end optimization.

\subsection{Sampling for the classification task} \label{exp-cls}
\textbf{Implementation details}.
We utilize the ModelNet40 \cite{wu-2015-cvpr} dataset that consists of 12311 manufactured CAD models in 40 categories.
Each CAD model are represented by 1024 points uniformly sampled from the mesh surface.
The train-test division follows the official setting.

We take the PointNet vanilla as the task network, which is identical to the comparison method \cite{lang-2020-cvpr}.
The employed task loss for classification is the widely used cross-entropy loss:
\begin{equation}\label{eq-task-cls}
\begin{aligned}
L_{task}(p,g) = - \sum_{k=1}^{N}{p_k \cdot \log(g_k)},\\
\end{aligned}
\end{equation}
where $p$ represents the softmax probability vector predicted by the classification network, $g$ is an one-hot vector converted from the ground-truth label, $p_k$ and $g_k$ represent the $k$-th element of $p$ and $g$, respectively.
Data augmentation is introduced to each batch of data during training, including random rotation and jittering.
The PointNet vanilla is trained by the Adam optimizer with a momentum of 0.9 and an initial learning rate of $1e-3$, which decays by 0.7 every 20 epochs until it reaches $1e-5$.
After 250 epochs of training with a batch size of 32, the task network finally gets 87.33\% accuracy with 1024 points at the inference time.

To train the AS-PD for classification, we set parameters: $\lambda=0.5$, $\alpha=10$ for the loss regularization.
The input size is fixed at 1024 points.
For the optimization in the first stage, we take the sample size of 32, batch size of 32, and 200 epochs.
Besides, we take a dynamic learning rate, which starts at $1e-3$ and decays by 0.7 every 20 epochs until it reaches $1e-5$.
In the second stage, we take another 300 epochs to retrain the complete sampler with sample size candidates of \{16, 32, 64, 128, 256, 512\}.
Other settings are identical to the first stage.

\begin{figure*}[!tbp] 
\centering 
\includegraphics[width=0.8\textwidth]{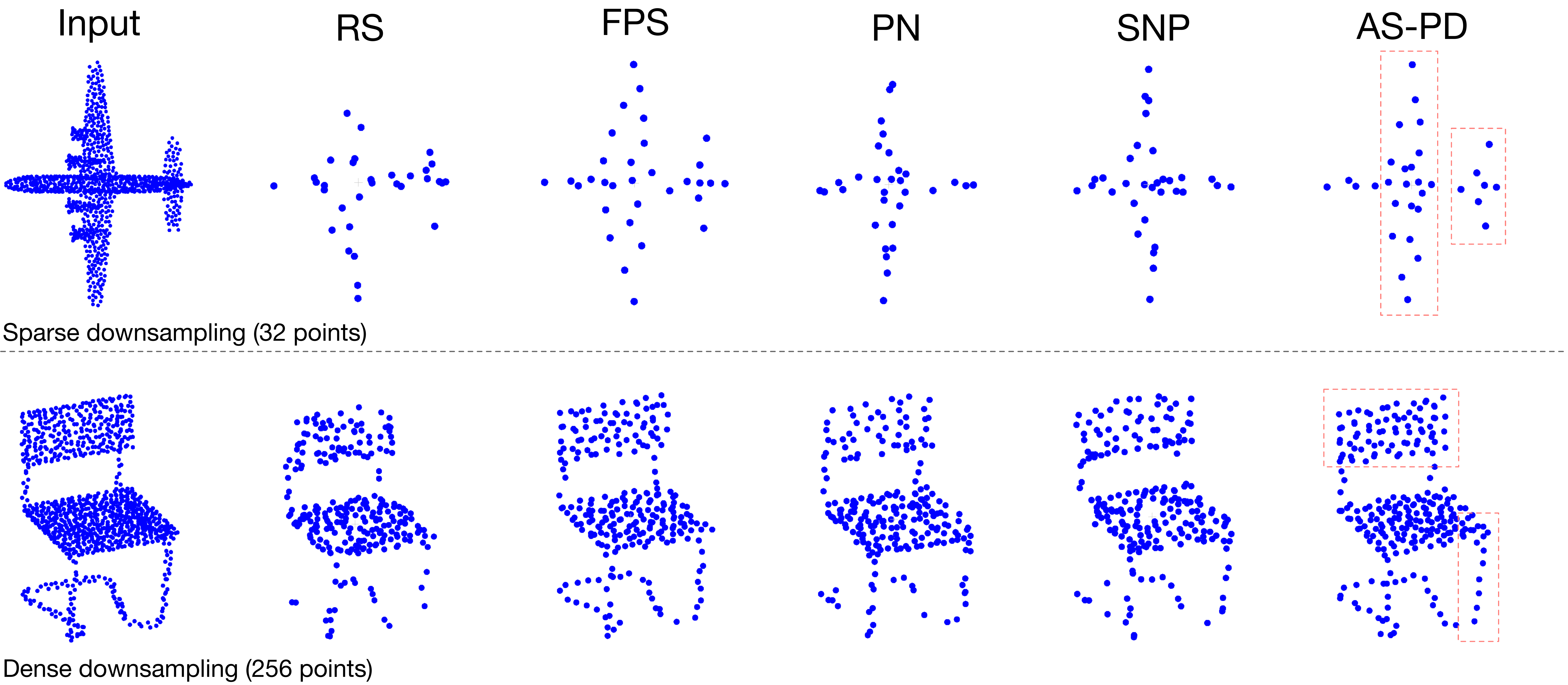} 
\caption{Visualization of sparse and dense sampled points for classification. 
The sample size for sparse and dense cases is set to 32 points and 256 points, respectively.} 
\label{Fig-cls-qualitative} 
\end{figure*}

\begin{figure}[tbp] 
\centering 
\subfigure[Acc of sampled points]{
    \begin{minipage}[t]{4cm}
    \includegraphics[scale=0.25]{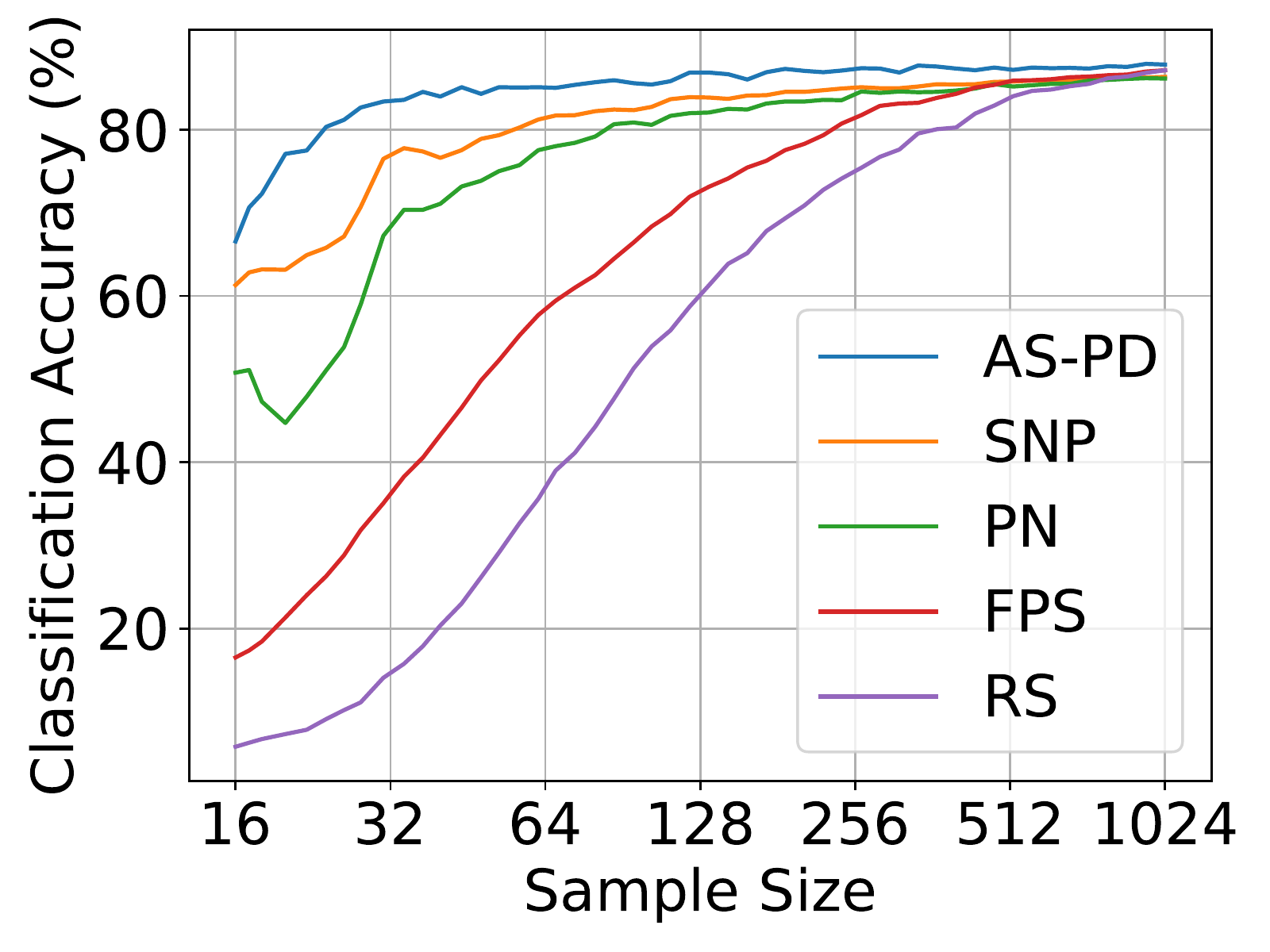}
    \label{Fig-cls-acc}
    \end{minipage}
}
\subfigure[HD between sampled points and input points]{
    \begin{minipage}[t]{4cm}
    \includegraphics[scale=0.25]{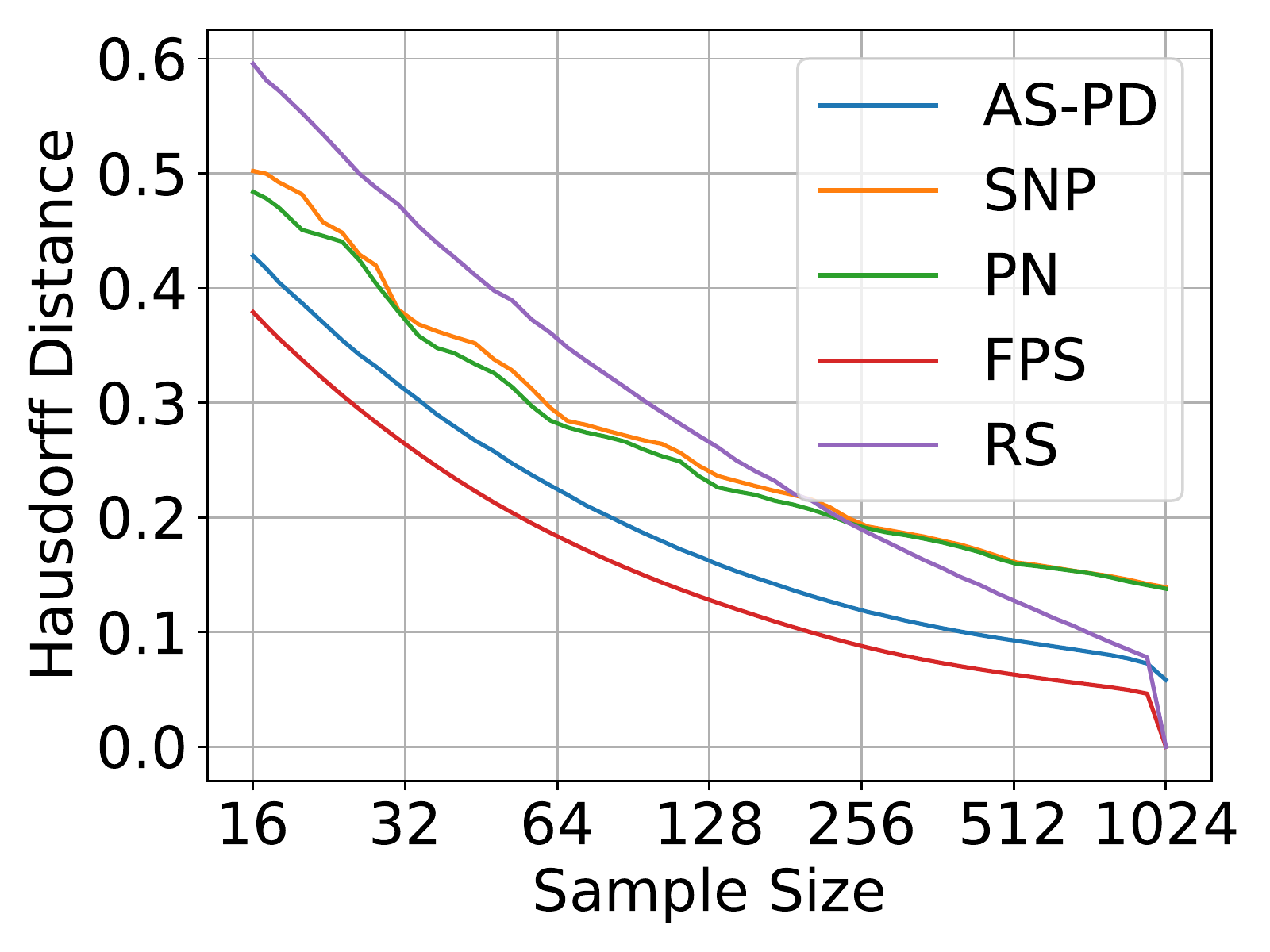}
    \label{Fig-cls-hd}
    \end{minipage}
}
\caption{Quantitative comparison of sampling for classification.} 
\label{Fig-quanti-cls} 
\end{figure}

\textbf{Qualitative analysis}.
To qualitatively compare the sampled result of the AS-PD with others, we visualize the sampled points of sparse (32 points) and dense (256 points) sample sizes, as illustrated in Fig. \ref{Fig-cls-qualitative}.
In Fig. \ref{Fig-cls-qualitative}, the proposed AS-PD does the best in preserving significant parts of the input.
For example, in the sparse downsampling case, the AS-PD well recovers the structures of the wing and tail of the airplane model, which are important for accurate recognition.
In the dense downsampling case, the AS-PD stores the details of the back and leg of the chair model, which are crucial ingredients of a chair model.

When it comes to other alternative samplers, the RS drops a lot of significant information in the input.
The FPS gives a fair sampled result by pursuing coverage of the input point cloud, but it falls short in accurately recovering the local geometry.
The PN and SNP only preserve the most important information for recognition while ignoring local details of the input, which limit their performance ceiling and transferability.

\textbf{Quantitative analysis}.
Fig. \ref{Fig-cls-acc} shows the Acc of sampled points by the RS, FPS, PN, SNP, and AS-PD.
The x-axis represents sample sizes from 16 to 1024 in a $log_2$ scale, and the y-axis represents the Acc.
From Fig. \ref{Fig-cls-acc}, the AS-PD surpasses all compared methods across the whole scope of sample sizes.
Furthermore, the Acc of the AS-PD increases more smoothly with the sample size compared to the PN and SNP, indicating better adaptability to unseen sample sizes.

Fig. \ref{Fig-cls-hd} indicates the HD between sampled points and input points.
From Fig. \ref{Fig-cls-hd}, the AS-PD achieves lower HD than any other methods except the FPS.
Considering the requirement of task-awareness, the HD of the AS-PD is slightly higher than the FPS which only considers the uniformity of distribution.
Even though we do not exploit the matching process which is used in PN and SNP to improve conformity, our AS-PD still achieve higher conformity than the PN and SNP.

\begin{figure*}[tbp] 
\centering 
\includegraphics[width=0.8\textwidth]{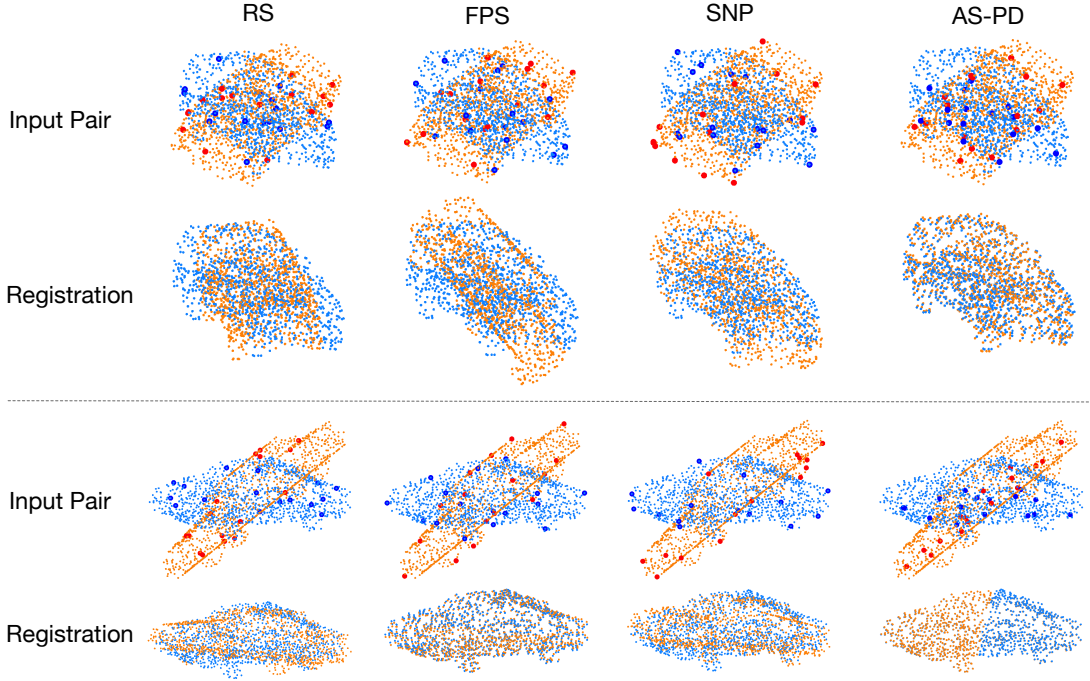} 
\caption{Visualization of registration using sampled points.
The sampled points of the template and source clouds are colored in blue and red, respectively.
The sample size is set to 16 points.} 
\label{Fig-regis-qualitative} 
\end{figure*}

\subsection{Sampling for the registration task} \label{exp-reg}
\textbf{Implementation details}.
We take the PCRNet \cite{sarode-2019-arXiv} as the task network for the point cloud registration.
We follow the experiment setting of Lang et al. \cite{lang-2020-cvpr} that takes 1024 points of the Car category in ModelNet40 \cite{wu-2015-cvpr} to train and evaluate.
To build template-source pairs for registration, the random rotation with Euler angles from -45$^\circ$ to 45$^\circ$ along three axes is implemented on each template point cloud.
Then, a source point cloud is obtained and corresponding rotation degrees are recorded.
Specifically, 4925 pairs of template and source point clouds are generated in the train split of dataset for training and another 100 pairs are generated in the test split for evaluation.
For training, we take a combined loss, consisting of an unsupervised chamfer distance loss and a supervised rotation error loss.
The rotation error is defined by the predicted rotation matrix $R_{pred}$ and the ground truth $R_{gt}$, where $R_{pred}$ is calculated based on the quaternion output of the PCRNet.
The task loss for registration can be given as:
\begin{equation}\label{eq-task-regist}
\begin{aligned}
L_{task} &= L_{unsup}(S,T) + L_{sup}(R_{pred},R_{gt}) \\
&= \frac{1}{|S|} \sum_{s \in S} \min_{t\in T} \|s-t\|^2 + \frac{1}{|T|} \sum_{t \in T} \min_{s\in S} \|t-s\|^2 \\ 
&+ \|R^{-1}_{pred} \cdot R_{gt} - I \|^2_{F}, \\
\end{aligned}
\end{equation}
where $I$ is a 3$\times$3 identity matrix and $\|\cdot\|_F$ represent the Frobenius norm.
After 500 epochs of training, the PCRNet can register source point clouds of 1024 points to the template one with a mean rotation error of 4.41$^\circ$.

To train the AS-PD for registration, we freeze the weights of PCRNet and train the AS-PD in the two-stage scheme.
First, we cut off the density attention sub-block and train the remaining part with 32 points pre-sampled by the FPS.
Then, we take the density attention sub-block back and load the pre-trained weights.
After that, we retrain the sampler with point sets pre-sampled to sizes randomly selected from \{16, 32, 64, 128, 256, 512\}, for each batch of data.
We take parameters: $\lambda=100, \alpha=10, \beta=1$ for the loss regularization, split data with batch size of 32, and set the input size to 1024 points.
The Adam optimizer with momentum of 0.9 is exploited for the optimization.
In the first stage, the learning rate exponentially decays from 1e-3 every 20 epochs until it reaches 1e-5 for 200 epochs.
Another 300 epochs are taken for the second stage training.

\textbf{Qualitative analysis}.
Fig. \ref{Fig-regis-qualitative} gives two examples of the registration using 32 sampled points by various samplers.
Note that the sampling process is simultaneously conducted on the source and template point clouds, both of which own the same shape and can be transformed to each other through rigid transformations.
Therefore, effective sampling demands not only the concentration on one point cloud, but also the consistency between the two for effective registration.
As shown in Fig. \ref{Fig-regis-qualitative}, the AS-PD samples consistent points across the template and source clouds.
From the first example in Fig. \ref{Fig-regis-qualitative}, we observe that uniform sampling like the FPS is not always appropriate, and may cause failure of registration.
Our AS-PD evolves from the FPS pre-sampled points, and put emphasis on some crucial regions, such as the bottom of the car, resulting in accurate registration with sparse sampled points (as low to 16 points).

\begin{figure}[tbp] 
\centering 
\subfigure[MRE of sampled points]{
    \begin{minipage}[t]{4cm}
    \includegraphics[scale=0.25]{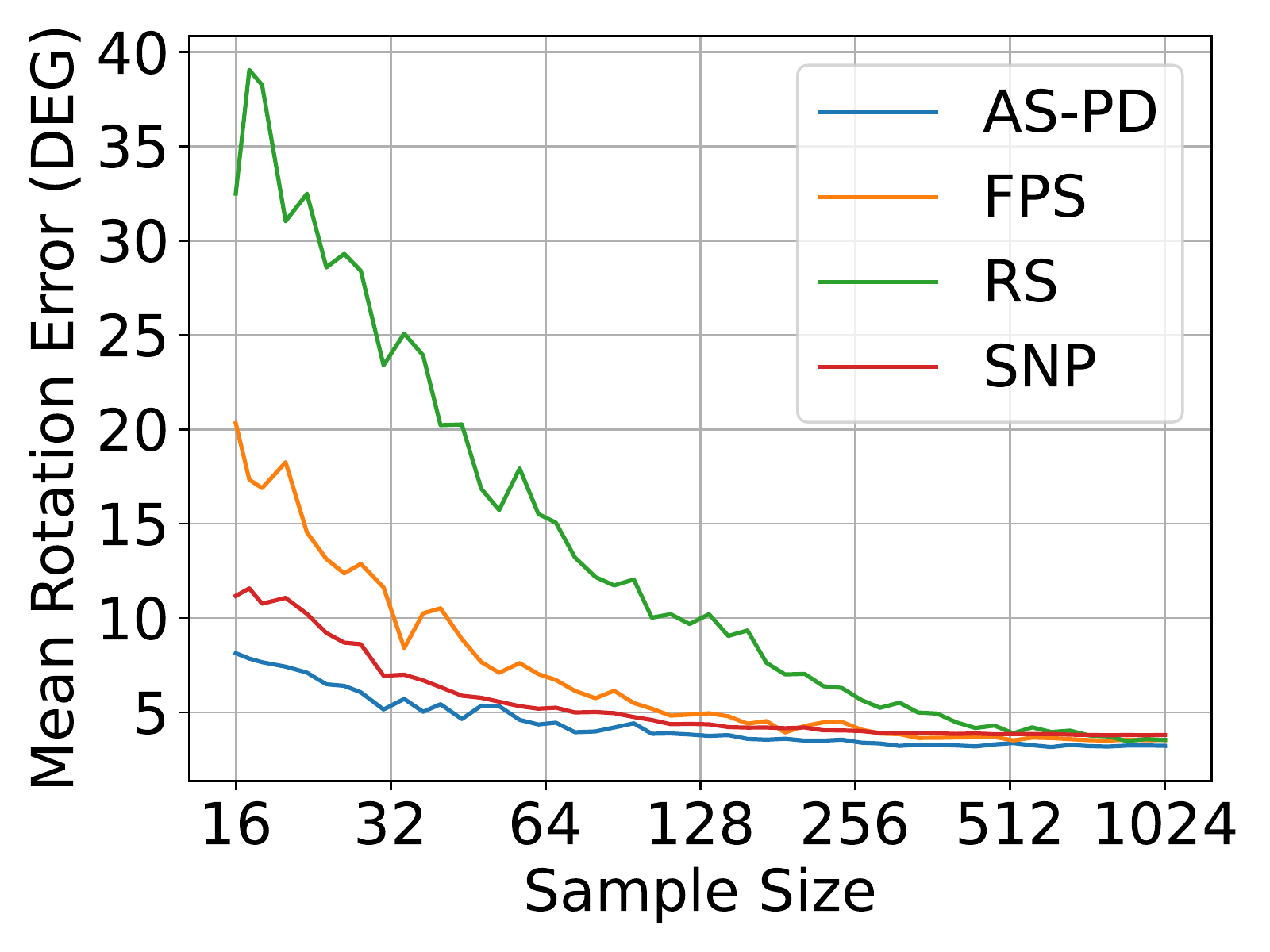}
    \label{Fig-regis-mre}
    \end{minipage}
}
\subfigure[HD between sampled points and input points]{
    \begin{minipage}[t]{4cm}
    \includegraphics[scale=0.25]{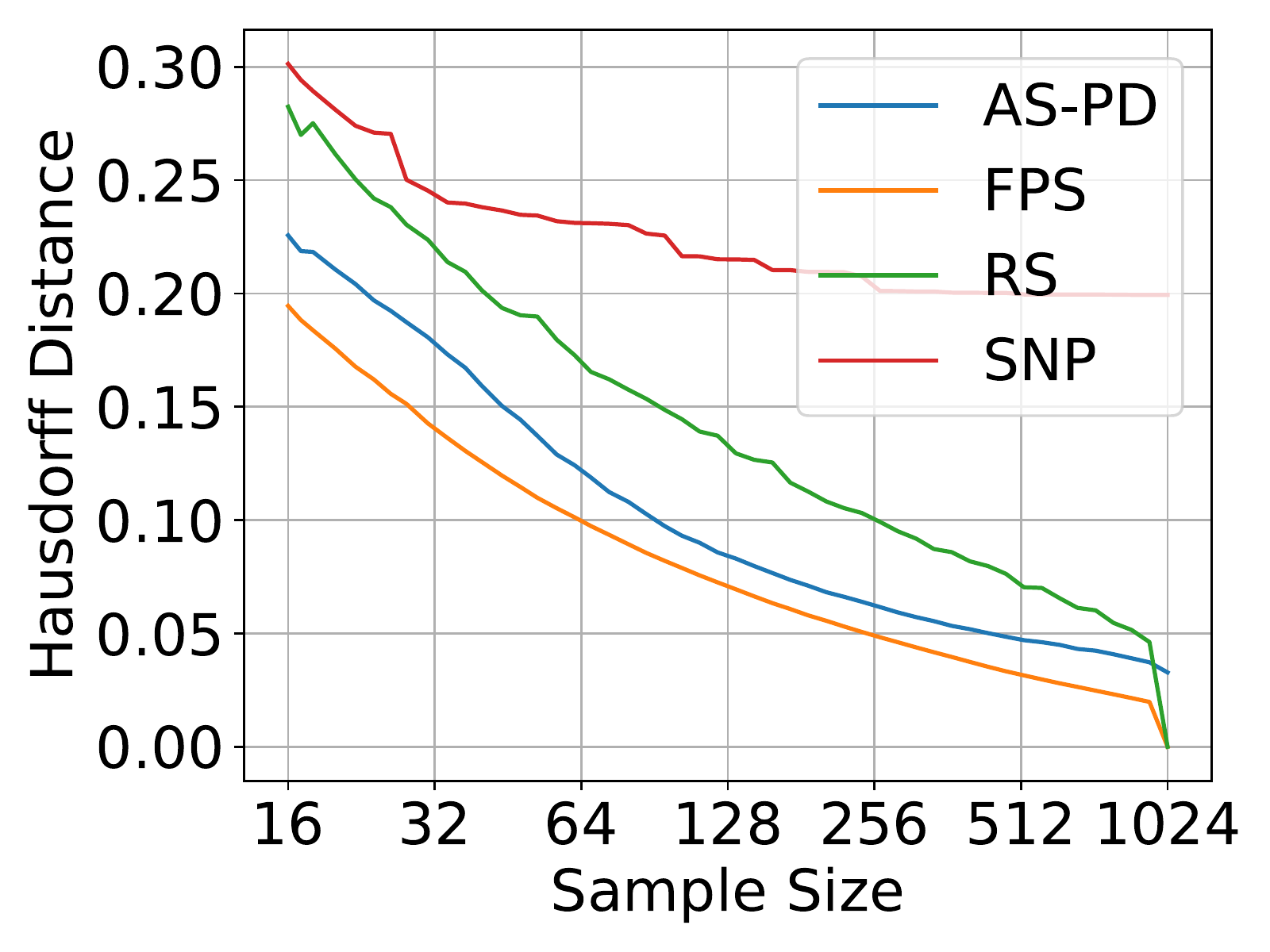}
    \label{Fig-regis-hd}
    \end{minipage}
}
\caption{Quantitative comparison of sampling for registration.} 
\label{Fig-quanti-regis} 
\end{figure}

\textbf{Quantitative analysis}.
The mean rotation error (MRE) is chosen as the metric to evaluate the performance of registration after sampling.
Fig. \ref{Fig-quanti-regis}  presents the quantitative comparison of various flexible sampling methods.
Fig. \ref{Fig-regis-mre} shows that the AS-PD achieves the lowest mean rotation error among all alternative samplers.
From Fig. \ref{Fig-regis-hd}, the HD of AS-PD is slightly higher than that of FPS, but significantly lower than that of SNP.

\begin{figure}[tbp] 
\centering 
\subfigure[32 sampled points]{
    \begin{minipage}[t]{4cm}
    \includegraphics[scale=0.25]{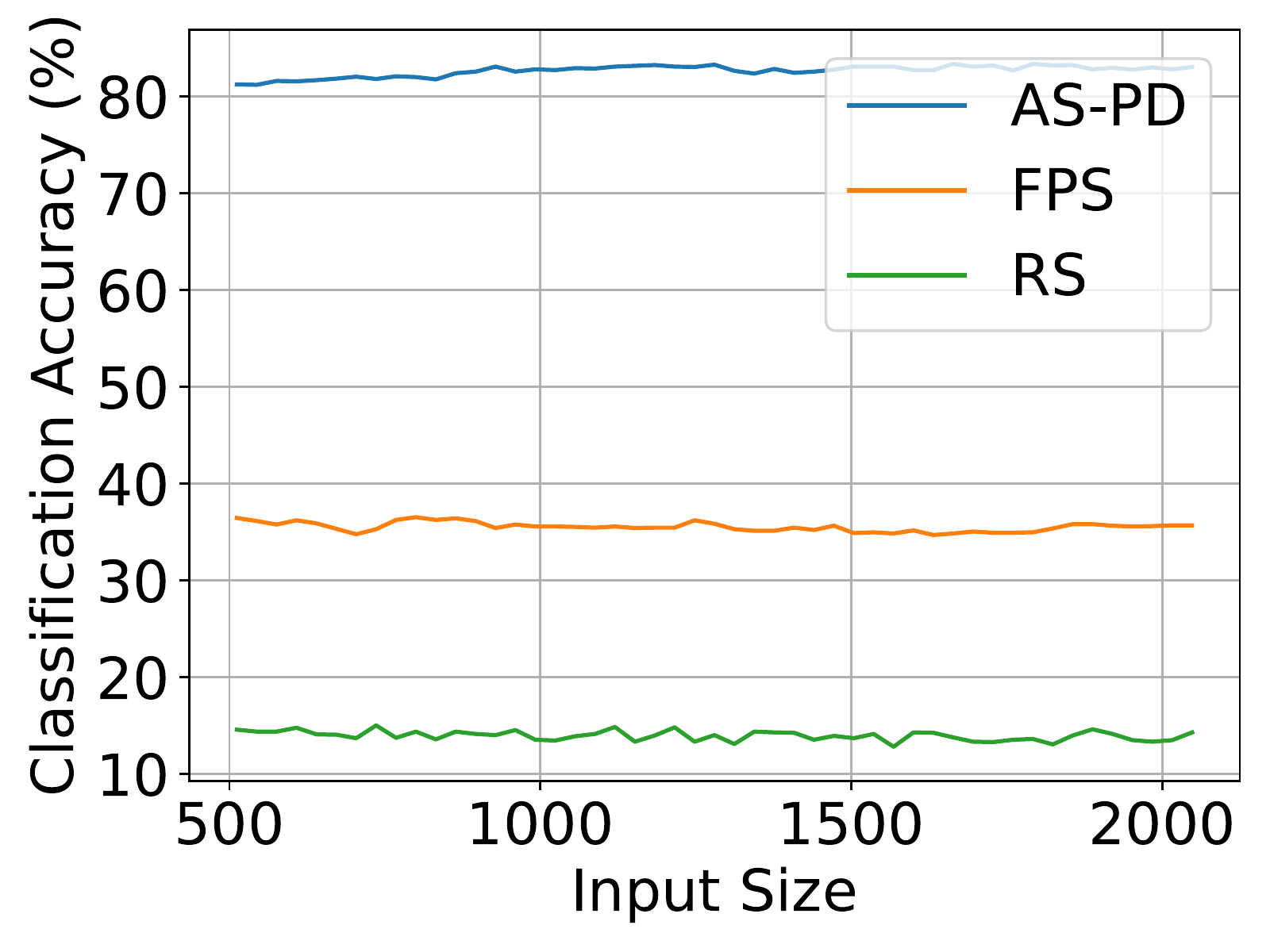}
    \label{Fig-vary-32}
    \end{minipage}
}
\subfigure[256 sampled points]{
    \begin{minipage}[t]{4cm}
    \includegraphics[scale=0.25]{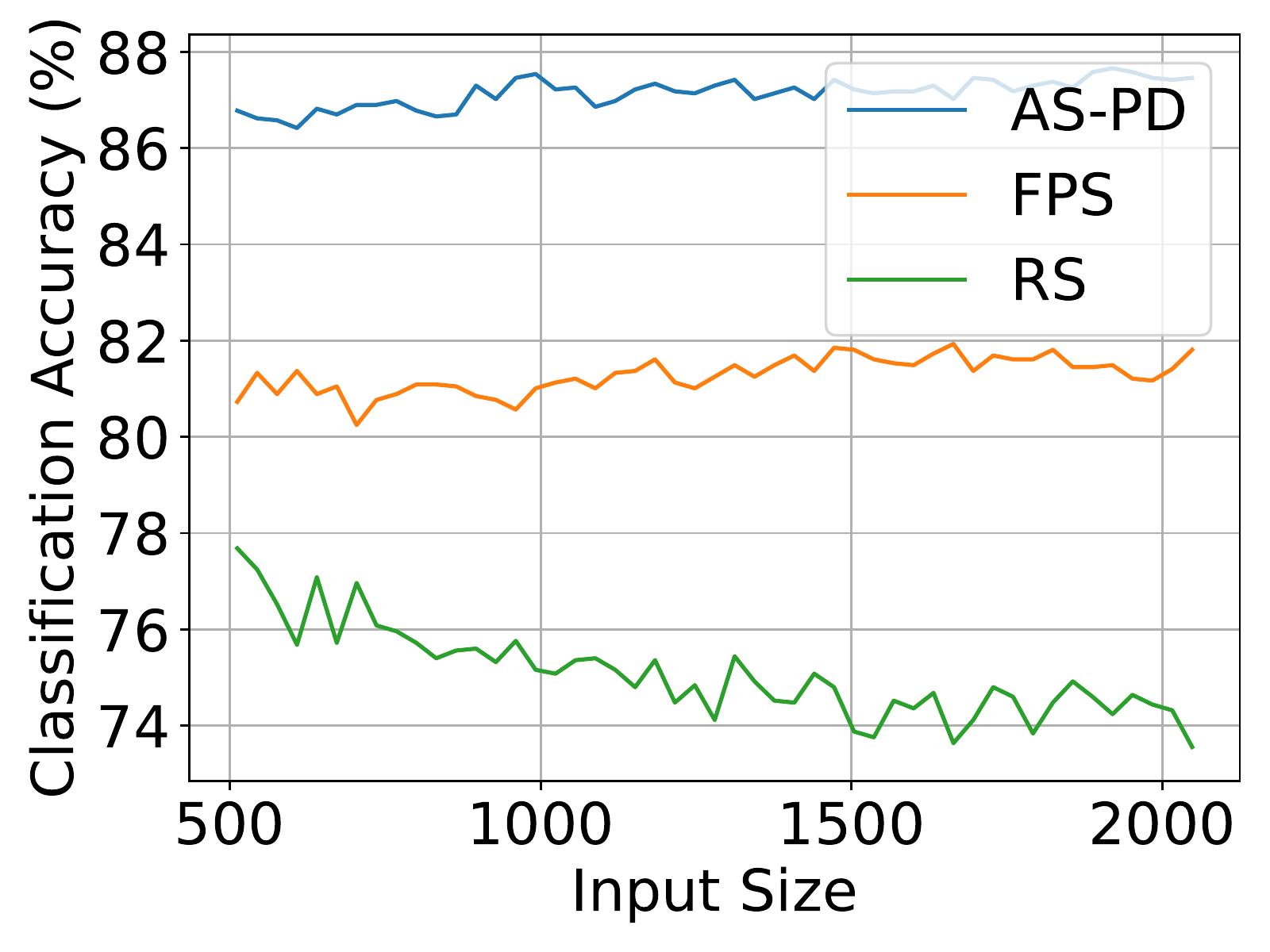}
    \label{Fig-vary-256}
    \end{minipage}
}
\caption{Classification accuracy of sampled points across different input sizes.} 
\label{Fig-varyinput} 
\end{figure}

\subsection{Flexibility across different input sizes} \label{exp-inputsize}
In the previous part, we have shown that the AS-PD could handle different sample sizes with one time training.
In this part, we conduct experiments on the classification task to check the flexibility of the AS-PD across different input sizes.

We include the FPS and RS as comparison methods.
Since the PN and SNP implement downsampling by reordering the input points, they are not suitable for different input sizes.
To train the AS-PD, we feed the sampler with different input sizes for each batch of data, ranging from 800 to 2000 points.
Specifically, the hyper-parameter $k$ in the DGCNN adaptively varies with the input size $n$, according to the prior below:
\begin{equation}\label{eq-varyinput}
k = k_0 * \frac{n}{n_0},
\end{equation}
where the $n_0$ and $k_0$ represent a couple of prior values of input size and neighborhood number.
Here we take $n_0=1024$ and $k_0=40$, which are used in the previous experiment setting.
Other experiment settings keep identical to the previous.

Fig. \ref{Fig-varyinput} depicts the classification accuracy of sampled points by the FPS, RS and AS-PD.
Without loss of generality, we choose the sample sizes of 32 and 256 points as examples.
The input size ranges from 512 to 2048 points for testing.
From Fig. \ref{Fig-varyinput}, the downstream performance of the AS-PD surpasses that of the FPS with a large margin.
More importantly, the effectiveness of the AS-PD is not impacted by the change of input sizes.
The fluctuation of the AS-PD across different input sizes is even smaller than that of heuristic methods (i.e., FPS and RS).

\begin{table}[tbp]
\normalsize
\centering
\caption{
Classification accuracy (\%) of sampled points on different task models.
Model 1 is the optimized task model of the sampler.
Model 2 and Model 3 are unseen task models, where Model 2 and Model 3 are different to Model 1 in parameters and structure, respectively.
$m$ represents the sample size.}
\setlength{\tabcolsep}{3mm}{ 
\begin{tabular}{ l | c c c }
\hline
 Methods & Model 1 & Model 2 & Model 3 \\
\hline
\multicolumn{4}{c}{$m$=32} \\
\hline
FPS & 35.6 & 33.6 & 30.5 \\
PN \cite{dovrat-2019-cvpr} & 68.7 & 37.8 & 28.6 \\
SNP \cite{lang-2020-cvpr} & 77.6 & 45.0 & 35.3 \\
AS-PD  & \textbf{83.0} & \textbf{64.4} & \textbf{53.5} \\
\hline
\multicolumn{4}{c}{$m$=256} \\
\hline
FPS & 81.1 & 81.9 & 82.5 \\
PN \cite{dovrat-2019-cvpr} & 84.3 & 82.3 & 81.6 \\
SNP \cite{lang-2020-cvpr} & 84.9 & 82.9 & 81.9 \\
AS-PD & \textbf{87.8} & \textbf{86.5} & \textbf{86.1} \\
\hline
\end{tabular} 
\label{Table-generality}
}
\end{table}

\subsection{Generality to unseen task models} \label{exp-generality}
In this part, we conduct experiments to check the generality of task-aware samplers to unseen task models of the same task.

The tested samplers contain learning-based methods (i.e., PN, SNP and AS-PD) and a task-agnostic method (i.e., FPS) as the baseline.
We take the classification as the downstream task without loss of generality.
For simplicity, we take the PointNet vanilla as the task model, which is used in previous experiments.
There are three different task models used for testing, termed model 1, model 2 and model 3.
Model 1 is the task model used to train samplers.
Model 2 shares the same structure as model 1, but the parameters are different.
Model 3 owns different structures and parameters to model 1.
Specifically, we formulate model 3 by changing the number of hidden layer nodes of model 1 from \{64, 64, 128, 1024\} to \{64, 256, 256, 1024\}.
Model 1 and model 2 are trained with different initialization methods to ensure that their parameters are different.
All task models are trained separately, and achieves similar classification accuracy on 1024 points (87.3\%, 87.2\% and 87.4\%).

Table \ref{Table-generality} reports the classification accuracy of sampled points on task models mentioned above.
We select the sample size of 32 and 256 as proxies for sparse and dense downsampling, respectively.
As for the sparse downsampling, all learning-based methods surpass the task-agnostic FPS with a large margin on the model 1 which has been seen during training.
When it comes to unseen models, the performance of learning-based methods drops
The PN and SNP drop over 30\% on the model 2, and 40\% on the model 3.
Specifically, the downstream performance of the PN (28.6\%) is even lower than that of the FPS (30.5\%) on the model 3.
In contrast, the AS-PD only drops 20\% to 30\% on unseen models, indicating generality to changes of the task model.
On the model 3, the AS-PD is still 23\% higher than the FPS.
As for the dense downsampling, the PN and SNP get worse performance than the FPS on the model 3, while the AS-PD keeps above 86\% on the two unseen models.

From the above observations we can conclude that the AS-PD is optimized to the task rather than some specific task model.

\subsection{Ablation study} \label{exp-ablation}
\begin{figure}[tbp] 
\centering 
\includegraphics[width=0.35\textwidth]{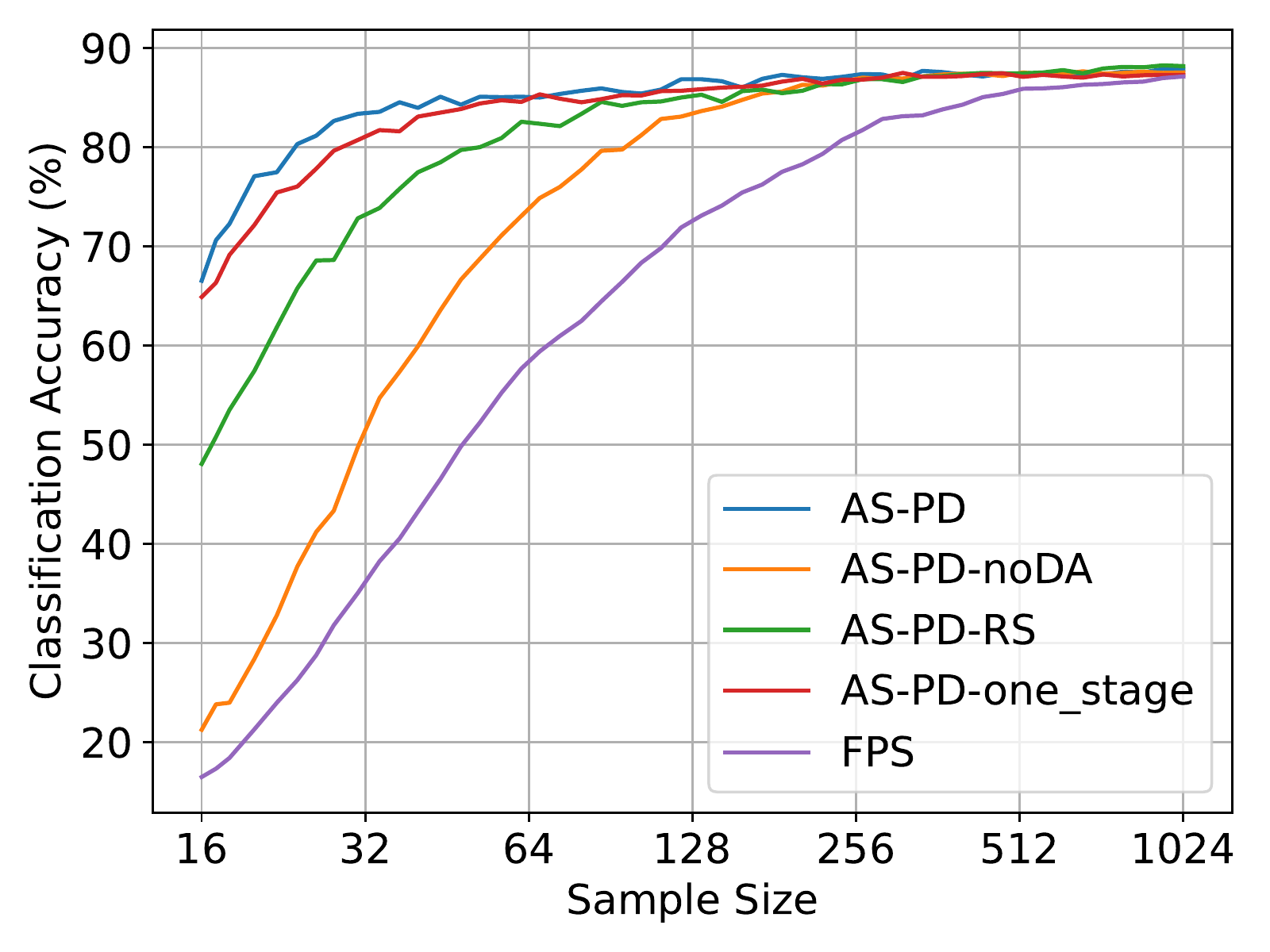} 
\caption{
Ablation studies on the AS-PD for classification.
AS-PD is the full pipeline of the proposed sampler.
AS-PD-noDA represents the degraded version by removing the density attention sub-block.
AS-PD-RS replaces the FPS pre-sampler in the AS-PD with the RS.
AS-PD-one\_stage has the same structure as the AS-PD, but is trained in one-stage scheme.
FPS is taken as a baseline method for comparison.
}
\label{Fig-cls-ablation} 
\end{figure}

To evaluate the importance of components in the pipeline of the AS-PD, we perform the AS-PD of different versions by removing or replacing some of them.
Fig. \ref{Fig-cls-ablation} shows the result of ablation studies for the classification task.

\textbf{Density attention sub-block.}
To check the effect of the density attention sub-block of the offset refining block, we conduct the ablation study by removing it and keeping the remaining experimental setup identical.
As shown in Fig. \ref{Fig-cls-ablation}, without the density attention sub-block, the Acc drops over 40\% in the worst case.
We interpret this phenomenon as the lack of sample size cues.
The density attention sub-block plays a key role in encoding the density of pre-sampled points, and making the sampler aware to different sample sizes.

\textbf{Pre-sampling}.
Pre-sampling is one of the core components of the AS-PD, which flexibly gives the initial task-agnostic sampled result.
Here we conduct the ablation study on the effect of different pre-sampling methods.
We choose two of the most widely used flexible heuristic methods, FPS and RS, as candidates for pre-sampling.

From Fig. \ref{Fig-cls-ablation}, the AS-PD equipped with the FPS pre-sampler achieves higher performance than the RS-based AS-PD at relative small sample sizes (smaller than 128 points).
The lack of the RS-based AS-PD in small sample sizes is mainly caused by the randomness of the RS pre-sampler.
As shown in Fig. \ref{Fig-cls-qualitative}, when doing sparse downsampling, the RS always fails to give reliable sampled result, and loses a lot of crucial information.
Even though, our offset refining block still cloud refine the RS pre-sampled result to a satisfied level.
For example, when downsampling 32 points, the RS-based AS-PD achieve 30\% improvement compared to the FPS.

However, when the sample size exceeds 128 points, the performance of the two AS-PD converges to the same level.
Note that the FPS is a time-expensive sampler when the sample size is large, while the RS has a significant advantage in speed regardless of the sample size \cite{hu-2021-tpami}.
Therefore, the choice is the pre-sampler is not set in stone.
When doing sparse downsampling, the FPS-based AS-PD is satisfied to balance the speed and performance.
When the sample size is large, the RS-based AS-PD is preferred to save time with little performance loss.

\textbf{Training scheme}.
To train the AS-PD, we design a two-stage training scheme.
Here we conduct ablation study on the impact of different training schemes, including the two-stage way and the simple one-stage way.
In the two-stage setting, we first cut off the density attention sub-block and train the sampler with a fixed sample size.
Then, we take the pre-trained weights to retrain the full version of the AS-PD.
In the one-stage setting, we remove the first stage in the two-stage setting and directly train the full AS-PD with a set of sample sizes.
Fig. \ref{Fig-cls-ablation} shows that the downstream performance is further improved by the two-stage training scheme.

\section{Conclusion}
We proposed an arbitrary-size downsampling framework for point clouds, which is optimized for downstream tasks.
The framework first samples point cloud to a given arbitrary smaller size in a heuristic way.
Then, it refines the pre-sampled set with point-based neural networks, getting the refined set optimized for downstream tasks.
Further, to let the sampler effectively learn across different sample sizes, we introduced the density attention sub-block to encode sample size cues.
Such a sample-to-refine approach realizes the task-aware sampling for arbitrary input size and sample size.
We demonstrated the effectiveness of our sampling framework with extensive experiments.
Compared to existing methods, our proposed sampler directly implements downsampling for arbitrary input size and sample size, and exhibits superior effectiveness and generality on downstream tasks.

\ifCLASSOPTIONcompsoc
  \section*{Acknowledgments}
\else
  \section*{Acknowledgment}
\fi

This work was supported by the National Key Research and Development Program of China under Grant 2022QY0102. The authors would like to acknowledge the helpful comments and kindly suggestions provided by anonymous referees.

\bibliography{my_reference}

\begin{IEEEbiography}[{\includegraphics[width=1in,height=1.25in,clip,keepaspectratio]
{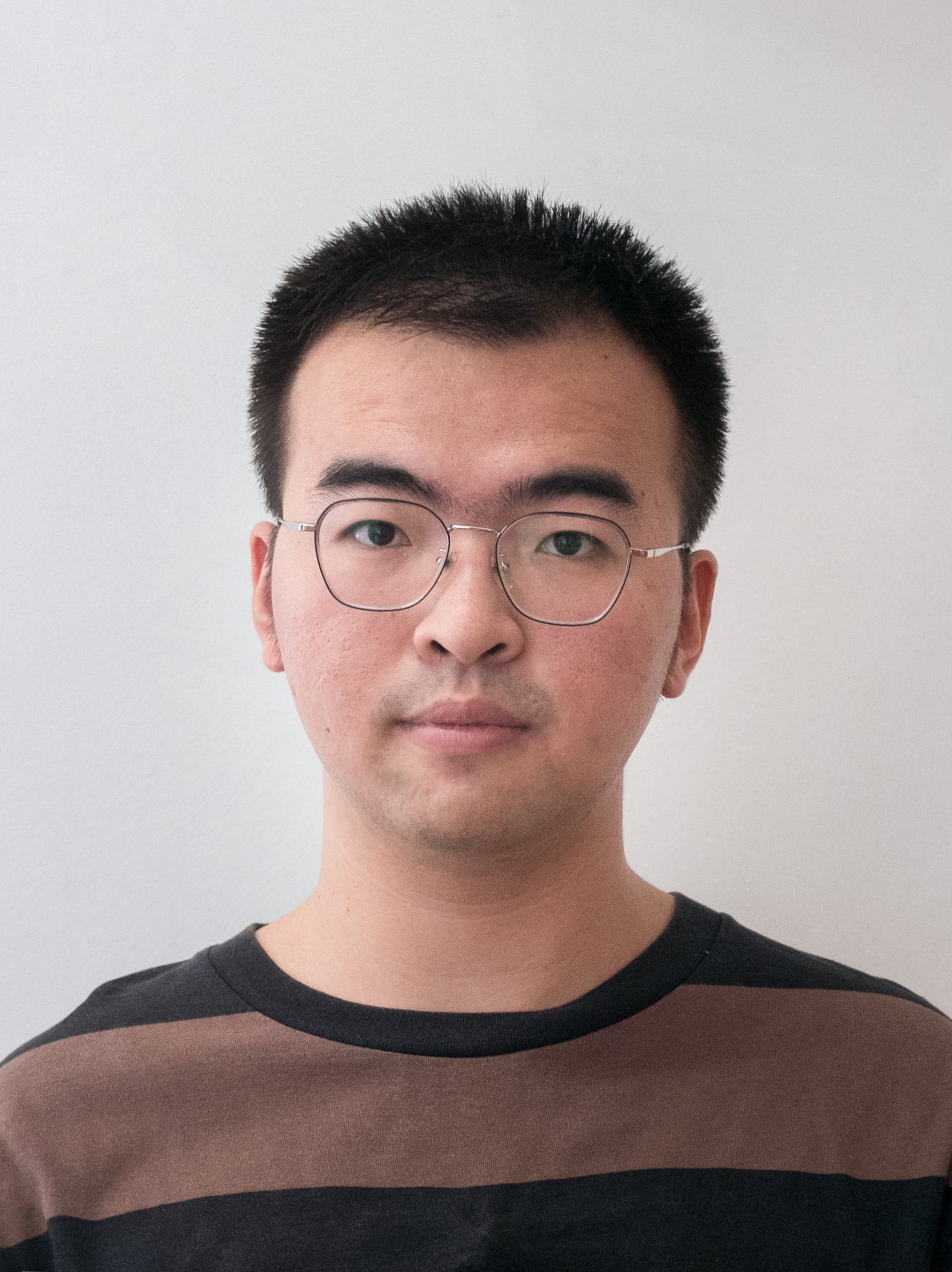}}]{Peng Zhang}
is currently pursuing his M.S. degree at the School of Automation, Nanjing University of Science and Technology, China. He received the bachelor's degree from the College of Electrical Engineering, Henan University of Technology, in 2020. His research interests include computer graphics, 3D computer vision and point cloud processing.
\end{IEEEbiography}

\begin{IEEEbiography}[{\includegraphics[width=1in,height=1.25in,clip,keepaspectratio]
{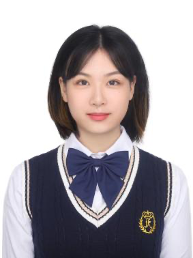}}]{Ruoyin Xie}
is pursuing her M.S. degree at School of Automation, Nanjing University of Science and Technology, China. She received the bachelor’s degree from the School of Energy and Power Engineering, Nanjing University of Science and Technology, in 2022. Her research interests include computer vision and point cloud processing.
\end{IEEEbiography}

\begin{IEEEbiography}[{\includegraphics[width=1in,height=1.25in,clip,keepaspectratio]
{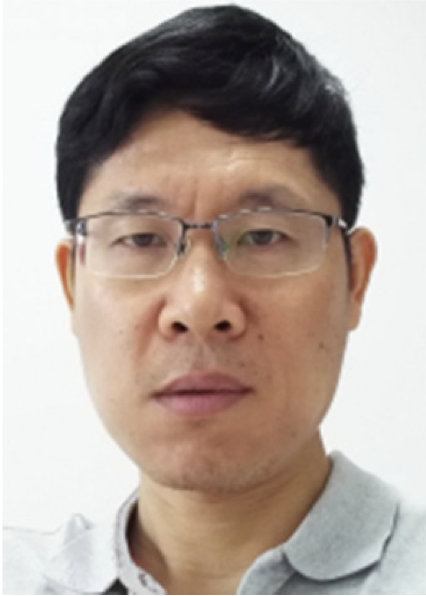}}]{Jinsheng Sun}
has been with the School of Automation, Nanjing University of Science and Technology (NUST), Nanjing, China, where he is currently a professor, since 1995. He received his B.S., M.S. and PhD of Control Science and Engineering from NUST in 1990, 1992 and 1995. From 2007 to 2009, he visited University of Melbourne as a Research Fellow at the Department of Electrical and Electronic Engineering. And in 2011, he was with the City University of Hong Kong as a senior Research Fellow. His research activity includes network congestion control, quality control and distributed control of multi-agent system.
\end{IEEEbiography}

\begin{IEEEbiography}[{\includegraphics[width=1in,height=1.25in,clip,keepaspectratio]
{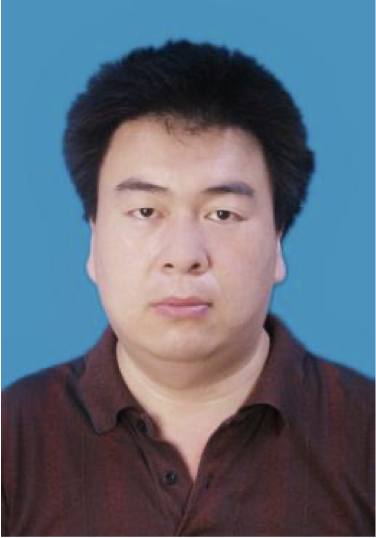}}]{Weiqing Li}
is currently an associate professor at the School of Computer Science and Engineering, Nanjing University of Science and Technology, China. He received the B.S. and Ph.D. degrees from the School of Computer Sciences and Engineering, Nanjing University of Science and Technology in 1997 and 2007, respectively. His current interests include computer graphics and virtual reality.
\end{IEEEbiography}


\begin{IEEEbiography}[{\includegraphics[width=1in,height=1.25in,clip,keepaspectratio]
{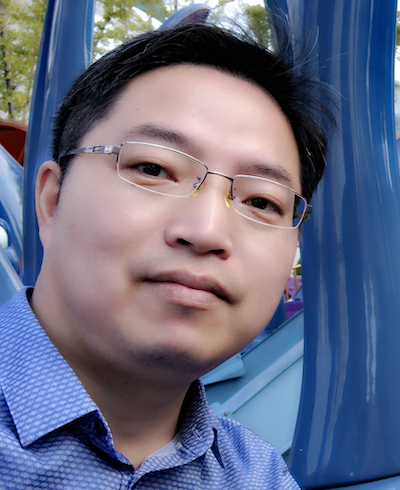}}]{Zhiyong Su}
is currently an associate professor at the School of Automation, Nanjing University of Science and Technology, China. He received the B.S. and M.S. degrees from the School of Computer Science and Technology, Nanjing University of Science and Technology in 2004 and 2006, respectively, and received the Ph.D. from the Institute of Computing Technology, Chinese Academy of Sciences in 2009. His current interests include computer graphics, computer vision, augmented reality, and machine learning.
\end{IEEEbiography}

\end{document}